\documentclass[conference]{IEEEtran}
\IEEEoverridecommandlockouts

\usepackage[T1]{fontenc}
\usepackage[utf8]{inputenc}
\usepackage{cite}
\usepackage{amsmath,amssymb}
\usepackage{graphicx}
\usepackage{booktabs}
\usepackage{array}
\usepackage{tabularx}
\usepackage{placeins}
\usepackage{url}
\usepackage{xcolor}
\usepackage{colortbl}
\usepackage{microtype}
\usepackage[hidelinks]{hyperref}
\graphicspath{{figures/}}
\title{HUMAID-NER: A Disaster Tweet Dataset for Joint Named Entity
Recognition and Event Classification via Uncertainty-Weighted
Multitask Learning}

\IEEEaftertitletext{%
\begin{center}\footnotesize
Published in: The Asian Bulletin of Big Data Management, 6(1), 138--152 (2026).\\
\url{https://doi.org/10.62019/zabvxd97}
\end{center}%
}

\makeatletter
\newcommand{\linebreakand}{%
  \end{@IEEEauthorhalign}\hfill\mbox{}\par\vspace{10pt}%
  \mbox{}\hfill\begin{@IEEEauthorhalign}%
}
\makeatother

\author{
\IEEEauthorblockN{Aijaz Ali}
\IEEEauthorblockA{\parbox[t]{0.30\textwidth}{\centering
\textit{Department of}\\
\textit{Software Engineering}\\
\textit{University of Sindh}\\
Jamshoro, Pakistan\\
{\small\nolinkurl{aijaz.laghari@students.usindh.edu.pk}}
}}
\and
\IEEEauthorblockN{Nazish Basir}
\IEEEauthorblockA{\parbox[t]{0.30\textwidth}{\centering
\textit{Department of}\\
\textit{Information Technology}\\
\textit{University of Sindh}\\
Jamshoro, Pakistan\\
{\small\nolinkurl{nazish.basir@usindh.edu.pk}}
}}
\and
\IEEEauthorblockN{Sarfaraz Nawaz}
\IEEEauthorblockA{\parbox[t]{0.30\textwidth}{\centering
\textit{Department of}\\
\textit{Software Engineering}\\
\textit{University of Sindh}\\
Jamshoro, Pakistan\\
{\small\nolinkurl{sarfaraz.mangi@students.usindh.edu.pk}}
}}
\linebreakand
\IEEEauthorblockN{Danish Nazir Arain}
\IEEEauthorblockA{\parbox[t]{0.36\textwidth}{\centering
\textit{Dr. A. H. S. Bukhari}\\
\textit{Postgraduate Centre of ICT}\\
\textit{University of Sindh}\\
Jamshoro, Pakistan\\
{\small\nolinkurl{danish.arain@usindh.edu.pk}}
}}
\and
\IEEEauthorblockN{Haris Ali}
\IEEEauthorblockA{\parbox[t]{0.36\textwidth}{\centering
\textit{Department of Software Engineering}\\
\textit{Mehran University of}\\
\textit{Engineering \& Technology}\\
Jamshoro, Pakistan\\
{\small\nolinkurl{harislag77@gmail.com}}
}}
}

\begin{document}

\maketitle
\pagestyle{plain}
\thispagestyle{plain}

% ══════════════════════════════════════════════════════════════════════════════
% ABSTRACT
% ══════════════════════════════════════════════════════════════════════════════
\begin{abstract}
Rapid extraction of structured information from social media is central
to effective humanitarian response, yet disaster tweet resources to date
offer only document-level category labels with no span-level entity
annotations. We address this gap with \textbf{HUMAID-NER}, the first
named entity recognition dataset built on the HumAID benchmark: 60{,}000
English disaster tweets annotated in BIO format across ten operationally
motivated entity types, including CASUALTY, DISPLACED, REQUEST,
RESOURCE, and RESCUE, yielding 21 entity classes and roughly 175{,}000
labelled entity spans. Annotations were produced through a reproducible
three-stage hybrid pipeline that combines a spaCy transformer backbone,
disaster-domain EntityRuler patterns, and structured regular expressions
with priority-based overlap resolution. We also propose a \textbf{joint
multitask learning framework} that performs disaster-specific NER and
humanitarian event classification through a single RoBERTa-large
encoder. A core difficulty in joint training is task-conflict: the NER
objective produces up to 2{,}688 token-level gradient signals per example
while classification contributes one, and under fixed task weights this
imbalance caused classification macro-F1 to fall 1.4 points across
epochs. Homoscedastic uncertainty weighting with learnable per-task
log-variance parameters resolves the conflict, paired with a two-stage
training schedule that freezes the lower 18 of 24 encoder layers in the
second stage to permit task-specific specialisation without eroding
shared representations. A controlled four-row ablation study isolates
each component's contribution. On the HUMAID-NER validation set, the
proposed system reaches NER span micro-F1 of \textbf{0.841} and
classification macro-F1 of \textbf{0.761} simultaneously; under this
setting, classification performance meets or exceeds dedicated
single-task RoBERTa-large classifiers on the same benchmark (0.730--0.750),
suggesting joint modelling introduces no classification trade-off while
adding complete entity extraction capability. A real-time web dashboard
demonstrates end-to-end deployment. Dataset, models, and pipeline code
are released to support reproducibility and future crisis informatics
research.
\end{abstract}


% ══════════════════════════════════════════════════════════════════════════════
\section{Introduction}
% ══════════════════════════════════════════════════════════════════════════════

When a major disaster strikes, affected communities turn to social media
almost immediately. Evacuation requests, casualty reports, and resource
needs appear within minutes of an event, producing a real-time
information stream that no structured reporting system can replicate
\cite{imran2015survey}. For humanitarian organisations, this creates a
critical operational problem: the sheer volume and noise make manual
monitoring impossible at crisis pace, yet buried within that stream is
exactly the actionable content that response coordinators need. NLP
systems that can filter, classify, and pull structured information out
of this stream are not an academic exercise; they are a practical
necessity for modern disaster response \cite{imran2015survey}.

Progress in classifying disaster-related social media by humanitarian
category has been considerable. The HumAID dataset \cite{alam2021humaid},
released by Alam et al.\ in 2021, provides roughly 77{,}000 human-labelled
English tweets from 19 major natural disaster events across ten
humanitarian classes, and transformer-based classifiers on this
benchmark have set strong performance standards \cite{alam2021humaid}.
The CrisisNLP corpora \cite{imran2016twitter} similarly showed, at an
earlier stage, that informational tweet categories can be reliably
identified through supervised deep learning. Classification alone,
however, does not fully satisfy operational requirements. Knowing that a
tweet belongs to the \textit{injured or dead people} category tells a
coordinator what type of message it is --- it does not say \textit{where}
casualties occurred, \textit{how many} are reported, or \textit{what
specific resources} are being sought. Those answers come from named
entity recognition (NER): direct extraction of typed spans from the
tweet text.

No existing dataset, to our knowledge, provides NER annotations designed
specifically for the disaster domain. Nor has any prior work trained a
single model to simultaneously handle disaster-specific NER and
disaster event classification. Social media NER resources
\cite{ritter2011ner}\cite{rijhwani2020soft} inherit generic entity
taxonomies from newswire benchmarks, omitting operationally critical
types such as CASUALTY, DISPLACED, RESOURCE, RESCUE, and REQUEST.
Running separate models for each task doubles inference cost and forfeits
the representational synergy that shared disaster vocabulary naturally
provides. The gap --- missing annotated data plus missing joint modelling
capability --- is the central problem this work takes on.

Joint multitask learning (MTL) with a shared encoder is, in principle,
well-suited to this setting \cite{liu2019mtdnn}. NER and classification,
however, differ fundamentally in gradient structure. NER generates dense
token-level supervision: 21 entity classes per token, producing up to
2{,}688 signals per example. Classification produces one sentence-level
signal per example. Under fixed-weight loss combination, the heavier NER
gradients systematically dominate optimisation, causing the
classification head to underfit beyond the early epochs. In our
controlled experiments, classification macro-F1 peaked at epoch three
and dropped by 1.4 percentage points by epoch fifteen --- a progressive,
structural failure rather than noise. This instability rules out
fixed-weight joint training for reliable deployment.

Our solution pairs homoscedastic uncertainty-based loss weighting
\cite{kendall2018uncertainty} --- which replaces static task weights
with learnable log-variance parameters --- with a two-stage training
procedure adapted from MT-DNN \cite{liu2019mtdnn}. Stage one builds
shared cross-task representations across all layers; stage two freezes
the bottom eighteen encoder layers and lets each task head specialise
using the remaining capacity. The backbone is RoBERTa-large
\cite{liu2019roberta}, chosen for its stronger pretraining recipe and
confirmed compatibility with TPU~v3-8. To support all of this, we extend
HumAID \cite{alam2021humaid} with disaster-specific NER annotations
across ten BIO-format entity types, producing HUMAID-NER --- the first
NER-annotated extension of this widely-cited benchmark.

The main contributions of this work are:
\begin{enumerate}
  \item We introduce \textbf{HUMAID-NER}, 60{,}000 tweets extending
    HumAID \cite{alam2021humaid} with BIO-format NER annotations across
    ten disaster-specific entity types, balanced across ten humanitarian
    classification labels, constituting the first NER-annotated resource
    built on the HumAID benchmark.
  \item We propose a \textbf{joint multitask framework} combining
    RoBERTa-large, Kendall uncertainty weighting
    \cite{kendall2018uncertainty}, and two-stage layer-freezing training
    \cite{liu2019mtdnn}. A controlled four-row ablation isolates each
    component's contribution and shows that uncertainty weighting
    prevents the classification degradation that fixed-weight training
    produces.
  \item We deploy the joint model as a \textbf{real-time web dashboard}
    for disaster response support, providing simultaneous entity
    extraction and event classification from live tweet input and
    demonstrating end-to-end applicability beyond academic evaluation.
\end{enumerate}

Section~II reviews related work. Section~III describes the dataset and
model. Section~IV presents experimental results, component analysis, and
the deployment dashboard. Section~V concludes.

% ══════════════════════════════════════════════════════════════════════════════
\section{Related Work}
% ══════════════════════════════════════════════════════════════════════════════

\subsection{Disaster Tweet Analysis}

Using social media as a real-time situational awareness source during
crises has been an active research area since the early 2010s. Imran
et al.\ \cite{imran2015survey} surveyed NLP methods for crisis messaging
and identified humanitarian information classification as the field's
central computational challenge. HumAID \cite{alam2021humaid} stands as
the most comprehensive English-language disaster tweet resource available
today, covering 19 disasters (2016--2019) with roughly 77{,}000 tweets
labelled across ten humanitarian categories; transformer-based systems on
this benchmark achieved macro-F1 of 0.70--0.75, establishing the
classification baseline this work builds on directly. Alam et al.\ also
extended the disaster tweet line with CrisisMMD \cite{alam2018crisisMMD},
a multimodal dataset pairing tweet text with images from seven 2017
events. That work, like HumAID and CrisisNLP \cite{imran2016twitter},
is limited to document-level categorical labels with no span-level entity
annotation. Broader work on disaster tweet categorisation for operational
use \cite{stowe2016disaster} and automated geo-event mapping from social
streams \cite{fan2020hybrid} further underlines the need for span-level
extraction alongside event classification.

\subsection{Named Entity Recognition on Social Media}

CoNLL-2003 \cite{tjong2003conll} formalised NER evaluation around four
entity types suited to newswire text, and those categories have dominated
benchmarks ever since. Ritter et al.\ \cite{ritter2011ner} documented
just how poorly standard NLP pipelines transfer to tweets: POS tagging
accuracy fell from 0.97 to 0.80, making tweet-specific models a
practical requirement. The gap narrowed considerably with transformer
fine-tuning \cite{devlin2019bert}, which reduced dependence on large
task-specific corpora through pretrained contextual representations ---
superseding the earlier contextual embedding approaches of Peters et al.\
\cite{peters2018elmo}. TweetNER7 \cite{ushio2022tweetner7}, a dedicated
Twitter NER benchmark with seven entity types across 11{,}382 English
tweets, reflects the community's continued investment in this problem.
Rijhwani et al.\ \cite{rijhwani2020soft} extended neural NER to
low-resource settings via soft gazetteers, incorporating cross-lingual
entity knowledge. HUMAID-NER draws on this principle: all ten entity
types are grounded in the operational vocabulary of humanitarian response
rather than inherited from general-purpose newswire categories.

\subsection{Multitask Learning for NLP}

Caruana \cite{caruana1997multitask} established the foundational argument
that joint training on related tasks improves generalisation, and Ruder
\cite{ruder2017overview} later surveyed the extensive NLP literature that
followed. For disaster tweets in particular, NER and event classification
share heavy vocabulary overlap --- \textit{collapsed, shelter, trapped,
evacuation} appear prominently in both task distributions --- which makes
joint training well-motivated on theoretical and empirical grounds.
MT-DNN \cite{liu2019mtdnn} showed this concretely: a single BERT encoder
jointly trained on sentence-level and token-level NLP tasks consistently
outperforms single-task fine-tuning, with lower encoder layers learning
universal linguistic features while upper layers encode task-specific
patterns. That layer-function insight directly motivates the two-stage
training procedure used here.

\subsection{Task Weighting in Multitask Learning}

Combining individual task losses into a single training objective is a
recurring challenge in MTL. Ruder \cite{ruder2017overview} identified
gradient imbalance as a primary driver of negative transfer: one task's
gradients simply overpower shared parameter updates. Kendall et al.\
\cite{kendall2018uncertainty} addressed this with homoscedastic
uncertainty weighting, giving each task a learnable log-variance
parameter that scales its loss contribution dynamically throughout
training. The log-variance parameterisation prevents collapse to trivial
solutions where $\sigma \rightarrow \infty$, and the approach transfers
cleanly from its original computer vision setting to the combination of
token-level NER and sentence-level classification studied here. PCGrad
\cite{yu2020gradient} offers an alternative that projects conflicting
gradients to eliminate destructive interference, but the approach
roughly doubles memory cost --- prohibitive at our training scale.
GradNorm \cite{chen2018gradnorm} provides a third option via dynamic
gradient magnitude scaling, though it similarly increases per-step
compute. Uncertainty weighting introduces only two scalar parameters with
negligible overhead, making it the practical choice for our setup. All
three methods are complementary, and a direct comparison on disaster-domain
MTL remains an open direction.

% ══════════════════════════════════════════════════════════════════════════════
\section{Methodology}
\label{sec:method}
% ══════════════════════════════════════════════════════════════════════════════

\subsection{Dataset: HUMAID-NER}

HumAID \cite{alam2021humaid} provides 77{,}637 English tweets spanning 19
major disasters (2016--2019), each labelled with one of ten humanitarian
categories but carrying no span-level entity annotations. We extended a
balanced 60{,}000-tweet subset with full BIO named entity annotations
across ten disaster-specific entity types to produce HUMAID-NER. The
dataset is partitioned into training (72\%), validation (14\%), and test
(14\%) splits, stratified by humanitarian category to preserve label
distribution. Table~\ref{tab:stats} reports the statistics.

\begin{table}[!t]
\renewcommand{\arraystretch}{1.2}
\caption{HUMAID-NER Dataset Statistics}
\label{tab:stats}
\centering
\footnotesize
\setlength{\tabcolsep}{3pt}
\begin{tabularx}{\columnwidth}{@{}l>{\centering\arraybackslash}X>{\centering\arraybackslash}X>{\centering\arraybackslash}X@{}}
\toprule
\textbf{Split} & \textbf{Tweets} & \textbf{Entity Spans} & \textbf{Avg/Tweet} \\
\midrule
Train (72\%)      & 43,200  & $\sim$126,000 & 2.91 \\
Validation (14\%) & 8,400   & $\sim$24,500  & 2.93 \\
Test (14\%)       & 8,400   & $\sim$24,500  & 2.92 \\
\midrule
\textbf{Total}    & \textbf{60,000} & $\sim$\textbf{175,000} & \textbf{2.92} \\
\bottomrule
\end{tabularx}
\end{table}

The entity taxonomy defines ten types rooted in the operational
vocabulary of humanitarian response rather than general-purpose newswire
categories \cite{tjong2003conll}: LOCATION, CASUALTY, DISPLACED,
REQUEST, RESOURCE, RESCUE, DISASTER\_TYPE, ORGANIZATION, PERSON, and
NUMBER, yielding 21 BIO classes (one O class and B-/I- prefixes for each
type). Table~\ref{tab:taxonomy} gives definitions and examples for each
type.

\begin{table*}[!t]
\renewcommand{\arraystretch}{1.25}
\caption{HUMAID-NER Entity Taxonomy}
\label{tab:taxonomy}
\centering
\small
\setlength{\tabcolsep}{8pt}
\begin{tabularx}{\textwidth}{@{}l>{\raggedright\arraybackslash}Xl@{}}
\toprule
\textbf{Type} & \textbf{Description} & \textbf{Example} \\
\midrule
LOCATION        & Geographical references         & ``Puerto Rico'' \\
DISASTER\_TYPE  & Named hazard or event type      & ``Hurricane Maria'' \\
CASUALTY        & Injury/fatality expressions     & ``17 dead'' \\
DISPLACED       & Evacuation/displacement counts  & ``1,000 evacuees'' \\
REQUEST         & Expressed need for aid          & ``need food'' \\
RESOURCE        & Available aid supplies          & ``water trucks'' \\
RESCUE          & Active emergency operations     & ``rescue teams'' \\
ORGANIZATION    & Named responding organisations  & ``Red Cross'' \\
PERSON          & Named individuals in tweets     & ``Mayor Cruz'' \\
NUMBER          & Standalone numerical values     & ``Day 3'' \\
\bottomrule
\end{tabularx}
\end{table*}

\begin{figure}[!t]
  \centering
  \includegraphics[width=\columnwidth]{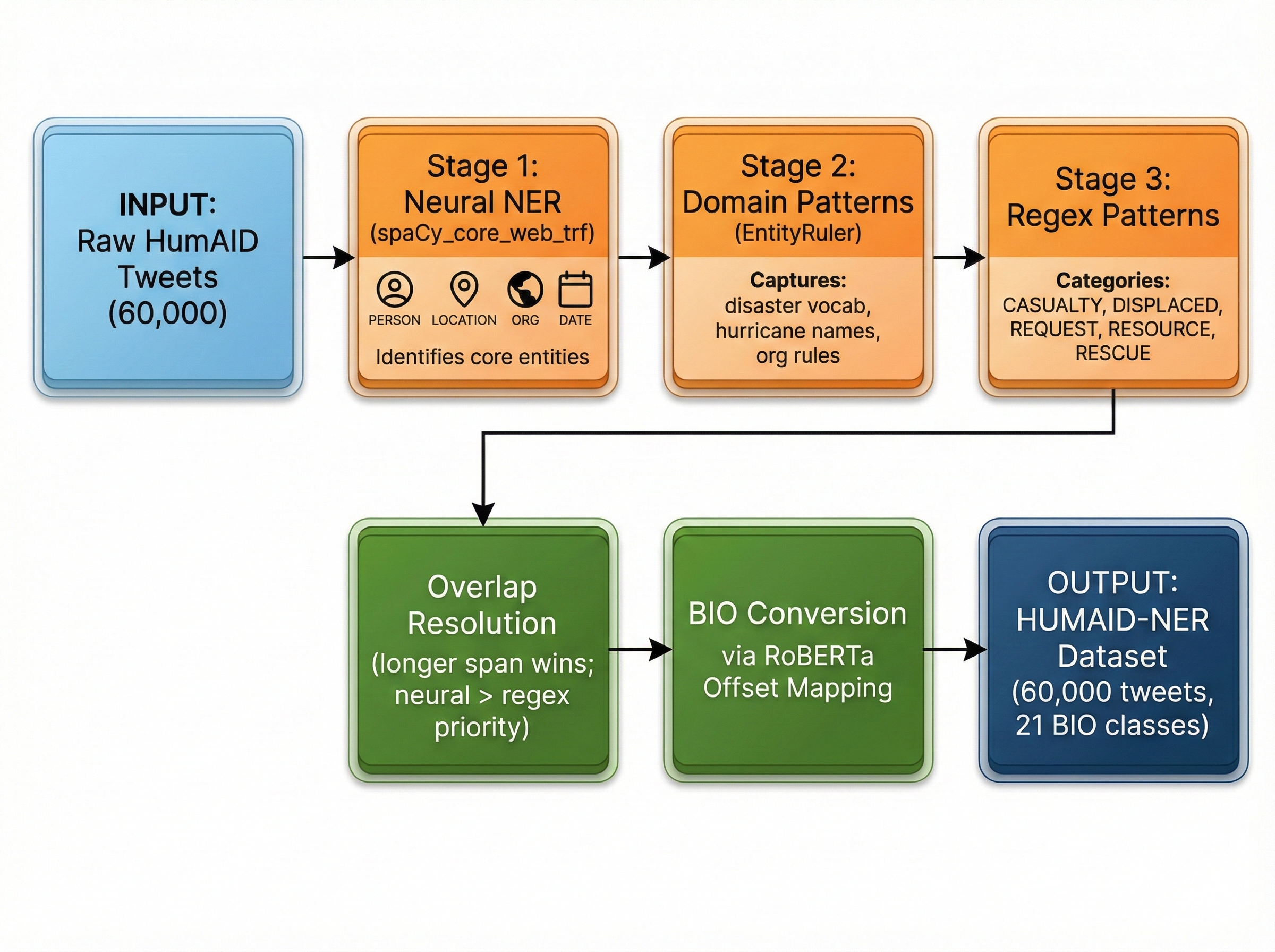}
  \caption{HUMAID-NER construction pipeline. Stage~1: spaCy transformer
  for base NER. Stage~2: disaster-domain EntityRuler patterns.
  Stage~3: regex for structured entity types. Overlap resolution and BIO
  conversion produce the final annotations.}
  \label{fig:pipeline}
\end{figure}

Manually annotating 60{,}000 tweets was not feasible, so we built a
three-stage hybrid auto-labelling pipeline, illustrated in
Fig.~\ref{fig:pipeline}. Stage~1 runs the spaCy \cite{honnibal2020spacy}
transformer model \texttt{en\_core\_web\_trf} to produce base entity
predictions for PERSON, LOCATION, ORGANIZATION, and DATE. Stage~2
passes the text through an EntityRuler component loaded with
domain-specific rules covering named disasters, humanitarian
organisations, and key disaster-vocabulary phrases. Stage~3 applies
regular expressions to capture the remaining structured types: CASUALTY,
DISPLACED, REQUEST, RESOURCE, and RESCUE. Where two spans compete, the
longer one is retained; for equal-length conflicts, neural predictions
take priority over regex matches. All annotations are then converted to
BIO format via offset mapping from the RoBERTa tokeniser, and
continuation subword tokens receive label $-100$ so they are excluded
from loss computation.

\subsection{Shared Encoder}

The backbone is RoBERTa-large \cite{liu2019roberta}, a transformer
\cite{vaswani2017attention} encoder with 24 hidden layers, 16 attention
heads, and hidden dimension $d{=}1{,}024$, totalling approximately 355M
parameters. We chose RoBERTa-large over BERT-large \cite{devlin2019bert}
for its stronger pretraining recipe, which uses dynamic masking,
full-sentence training objectives, and a 50{,}265-token byte-level BPE
vocabulary. DeBERTa-v3 \cite{he2021deberta} was also evaluated but
rejected after incompatible XLA gather operations on TPU~v3-8 hardware
prevented stable training. For a tweet tokenised to $T$ subword tokens,
the encoder outputs context-sensitive representations
$\mathbf{H} = \{h_1, \ldots, h_T\}$ where $h_i \in \mathbb{R}^{1024}$.

\subsection{Task-Specific Heads}

Two task heads branch from the shared encoder. The \textbf{NER head}
applies a linear projection to each token representation and produces
logits over $C{=}21$ entity classes; continuation subword tokens are
masked at $-100$ and excluded from loss computation. The
\textbf{classification head} takes a different approach: it concatenates
the [CLS] embedding $h_1$ with the mean of all non-padding token
representations, $\bar{h} = (1/T')\sum h_i$, to form a
2{,}048-dimensional input vector. This dual-pooling strategy
\cite{liu2019mtdnn} captures both the compressed global summary and
distributed token-level context. The concatenated vector passes through
a linear layer and dropout ($p{=}0.1$) to produce logits over $K{=}10$
humanitarian categories.

\begin{figure}[!t]
  \centering
  \includegraphics[width=\columnwidth]{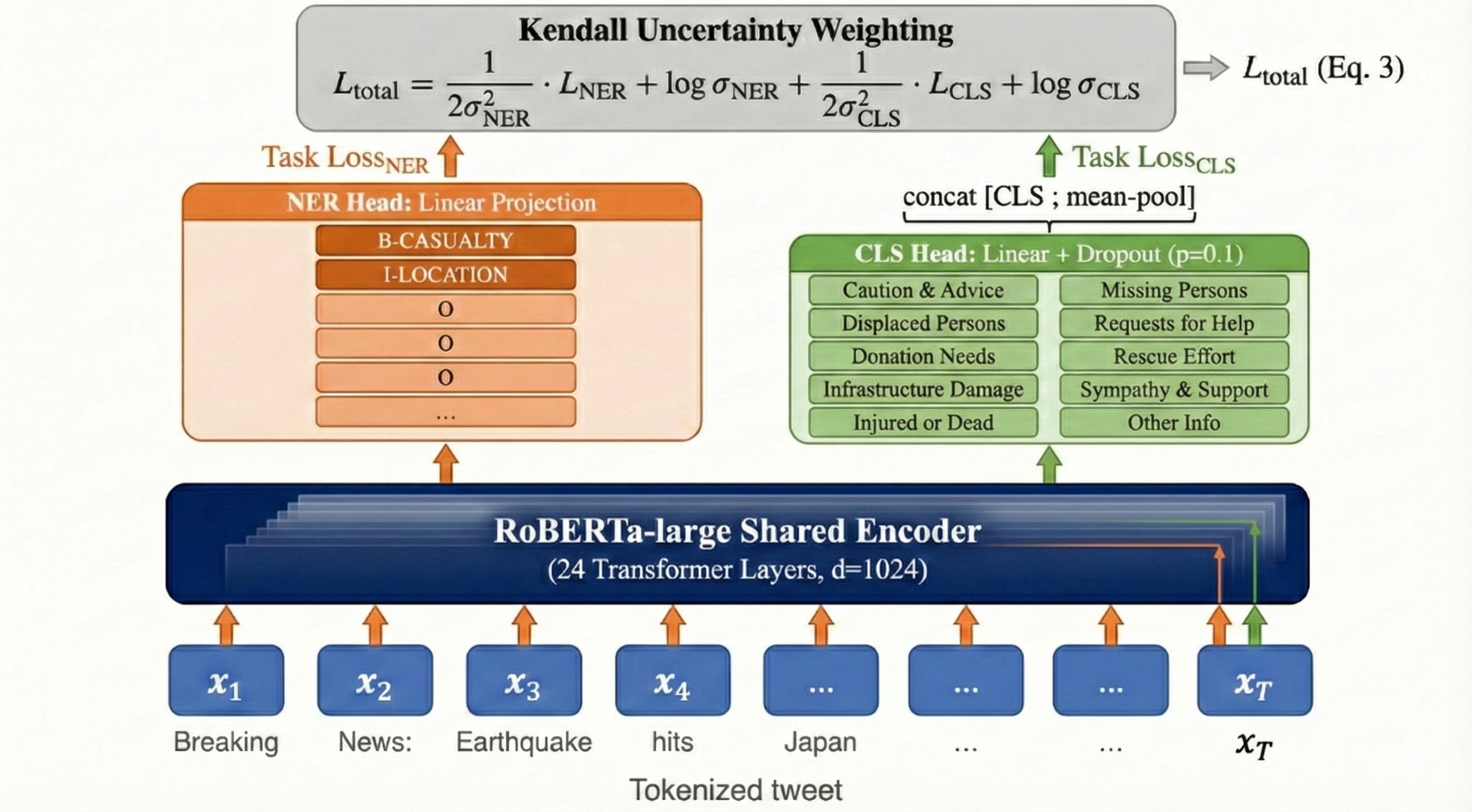}
  \caption{Model architecture. The shared RoBERTa-large encoder feeds
  two task heads. The NER head performs token-level classification over
  21 BIO entity classes. The CLS head concatenates [CLS] and mean-pooled
  representations before classifying into 10 humanitarian categories.
  Learnable uncertainty parameters $\sigma_\text{NER}$ and
  $\sigma_\text{CLS}$ weight the combined loss.}
  \label{fig:arch}
\end{figure}

\subsection{Uncertainty-Weighted Multitask Objective}

Both task losses are standard cross-entropy objectives. The NER loss
averages across all non-padding token positions:
\begin{equation}
  \mathcal{L}_\text{NER} = -\frac{1}{N}\sum_i\sum_c
    y_{ic}\log p_{ic}
  \label{eq:ner}
\end{equation}
where $N$ is the non-padding token count, $y_{ic}\in\{0,1\}$ is the
ground-truth indicator, and $p_{ic}$ is the predicted probability. The
classification loss is:
\begin{equation}
  \mathcal{L}_\text{CLS} = -\sum_k y_k \log p_k
  \label{eq:cls}
\end{equation}

Fixed-weight combination is problematic here because
$\mathcal{L}_\text{NER}$ aggregates up to $128{\times}21{=}2{,}688$
supervision signals per example while $\mathcal{L}_\text{CLS}$
contributes just one. We therefore adopt the homoscedastic uncertainty
weighting of Kendall et al.\ \cite{kendall2018uncertainty}, which derives
the joint objective from a probabilistic log-likelihood perspective:
\begin{equation}
  \mathcal{L}_\text{total} =
  \sum_{t\in\{\mathrm{NER},\,\mathrm{CLS}\}}
  \left(\frac{\mathcal{L}_t}{2\sigma_t^2}+\log\sigma_t\right)
  \label{eq:total}
\end{equation}
The $1/(2\sigma_i^2)$ terms scale each task loss inversely with
uncertainty, automatically down-weighting whichever task currently
dominates. The $\log\sigma_i$ regularisation terms block trivial
solutions where $\sigma \rightarrow \infty$. For numerical stability we
reparameterise $s_i{=}\log(\sigma_i^2)$, giving $1/(2\sigma_i^2) =
e^{-s_i}/2$ and $\log\sigma_i = s_i/2$. Substituting yields the
optimised form used in practice:
\begin{equation}
  \mathcal{L}_\text{total} =
  \frac{1}{2}\sum_{t\in\{\mathrm{NER},\,\mathrm{CLS}\}}
  \left(e^{-s_t}\mathcal{L}_t+s_t\right)
  \label{eq:totalopt}
\end{equation}
Both $s_i$ are initialised to 0 (i.e., $\sigma_i{=}1$, equal initial
weighting) and converge to approximately 1.26 by epoch~15. Since every
term in Eq.~\eqref{eq:totalopt} is non-negative for $s_i{>}0$, the
total loss stays positive throughout training. Both
$\mathcal{L}_\text{NER}$ and $\mathcal{L}_\text{CLS}$ decrease
monotonically, confirming that neither task is neglected as the
uncertainty parameters adapt.

\subsection{Two-Stage Training Procedure}

Training runs in two sequential stages, motivated by the layer-function
analysis in Liu et al.\ \cite{liu2019mtdnn}: lower encoder layers learn
universal features shared across tasks, while upper layers encode
task-specific patterns.

\textbf{Stage~1} (epochs 1--7): All 24 encoder layers, both task heads,
and the two uncertainty parameters are trained jointly using
Eq.~\eqref{eq:total}. The learning rate warms up linearly over 1{,}125
steps to a peak of $2{\times}10^{-5}$, then decays linearly back to
zero. The stage boundary at epoch~7 was set empirically by tracking
the validation combined score, which plateaued between epochs~6 and~8.

\textbf{Stage~2} (epochs 8--15): Layers 1--18 are frozen via
\texttt{requires\_grad = False}, leaving only layers 19--24, both task
heads, and the uncertainty scalars to receive gradient updates. This
reduces active parameters from 355M to roughly 47M. Stage~2 restarts
with the same peak learning rate and a fresh warmup over 450 steps.

\begin{figure}[!t]
  \centering
  \includegraphics[width=\columnwidth]{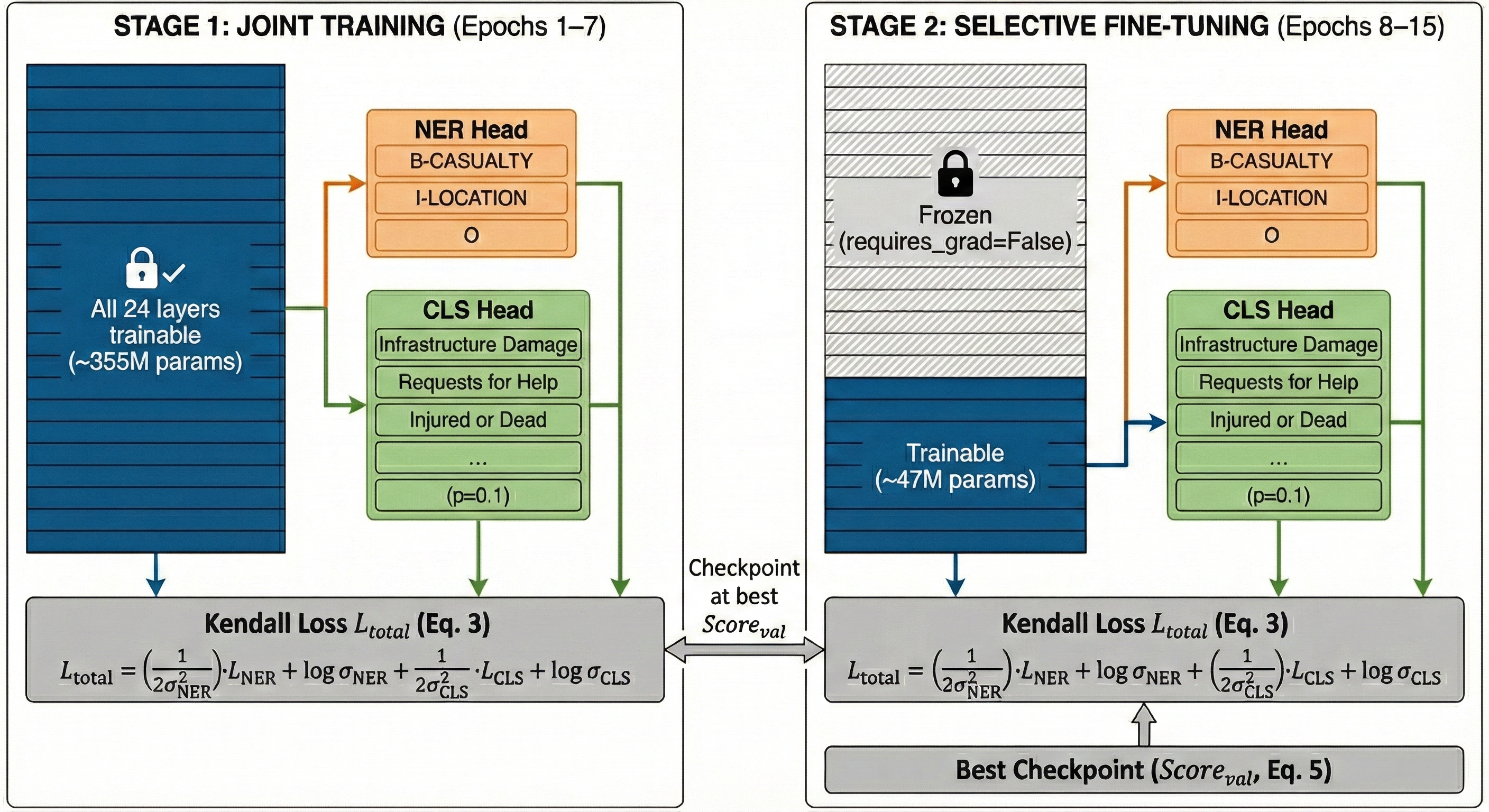}
  \caption{Two-stage training procedure. Stage~1 (epochs 1--7) trains
  all layers jointly. Stage~2 (epochs 8--15) freezes layers 1--18 and
  fine-tunes only the upper six layers and task heads, reducing active
  parameters from 355M to 47M.}
  \label{fig:stages}
\end{figure}

\subsection{Optimisation and Evaluation}

All models use AdamW \cite{loshchilov2019adamw} ($\beta_1{=}0.9$,
$\beta_2{=}0.999$, $\lambda{=}0.01$) with gradient clipping at norm
1.0, batch size 8, and maximum sequence length 128 tokens, run on a
TPU~v3-8 with PyTorch~2.6 \cite{paszke2019pytorch} and HuggingFace
Transformers \cite{wolf2020transformers}. NER performance is measured
with entity-level span micro-F1 via seqeval \cite{nakayama2018seqeval}:
a predicted span counts as correct only when both the boundary and
entity type exactly match the gold annotation. Classification is measured
with macro-averaged F1 across the ten humanitarian categories. During
training, checkpoint selection relies on the balanced combined score:
\begin{equation}
  \text{Score}_\text{val} = 0.5 \times F1_\text{NER} +
  0.5 \times F1_\text{CLS}
  \label{eq:score}
\end{equation}
Configuration selection across Rows A--D follows a \textit{classification-first
deployment policy}: because downstream humanitarian response routing
depends directly on the event category label, CLS macro-F1 is the
primary criterion and combined score breaks ties. This policy is
declared here and applied consistently throughout
Section~\ref{sec:results}.

% ══════════════════════════════════════════════════════════════════════════════
\section{Results and Demonstration}
\label{sec:results}
% ══════════════════════════════════════════════════════════════════════════════

All results are reported on the held-out validation split (8{,}400 tweets)
using the best checkpoint selected by Eq.~\eqref{eq:score}. The figures
in this section were generated directly from training logs and reflect
empirically observed values.

\subsection{Ablation Study}

\begin{figure*}[!t]
  \centering
  \includegraphics[width=0.7\textwidth]{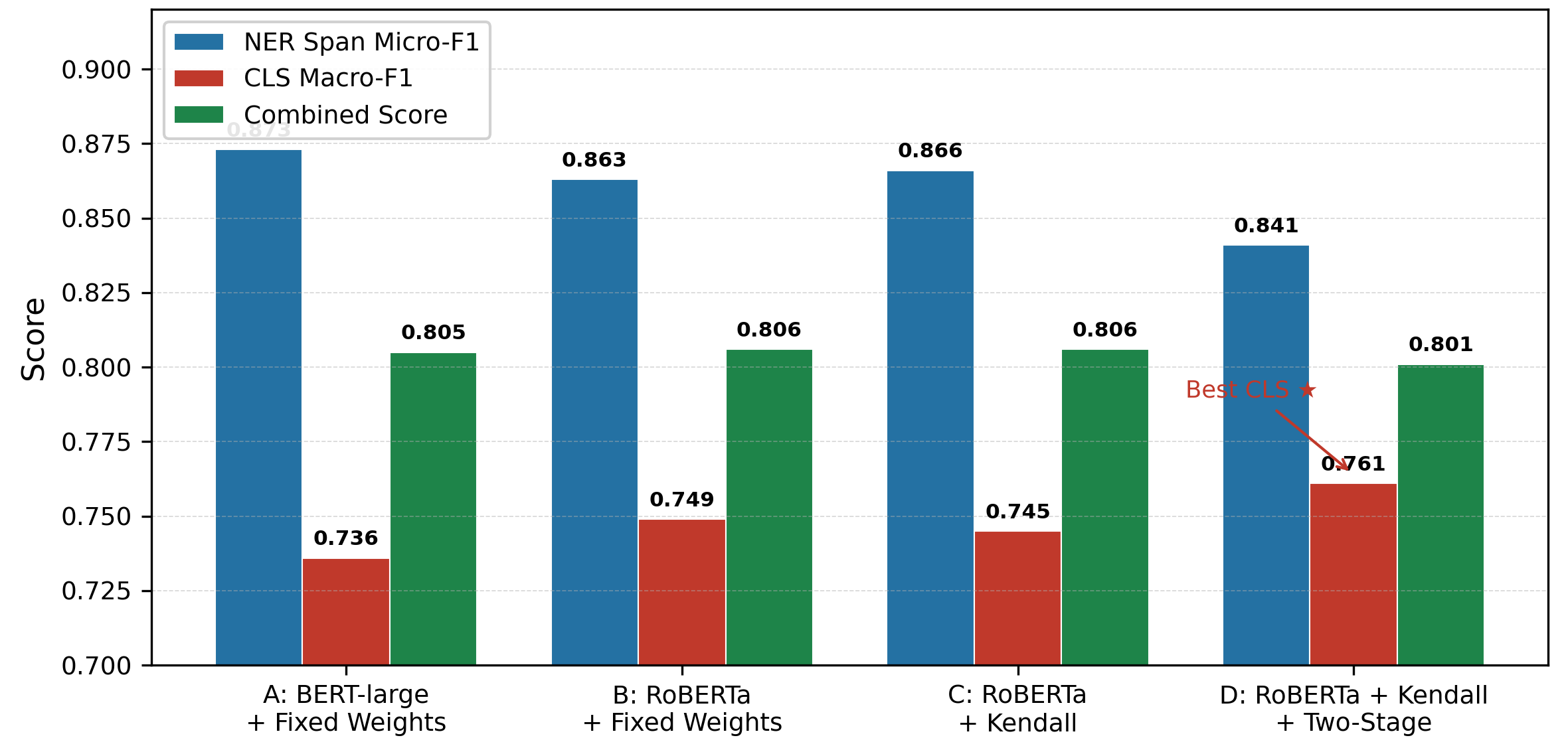}
  \caption{Ablation scores for Rows A--D. NER span micro-F1 (blue),
  CLS macro-F1 (red), combined score (green). Row~A achieves the highest
  NER (0.873); Row~D achieves the highest CLS (0.761), the operationally
  critical metric for deployment.}
  \label{fig:ablation}
\end{figure*}

Table~\ref{tab:ablation} presents the four-row controlled ablation. Each
row introduces exactly one new component: Row~A provides the BERT-large
baseline under fixed weights; Row~B swaps in RoBERTa-large; Row~C adds
Kendall uncertainty weighting; Row~D adds two-stage freezing to form the
complete proposed system. Fig.~\ref{fig:ablation} displays all three
metrics side by side.

\begin{table*}[!t]
\renewcommand{\arraystretch}{1.2}
\caption{Ablation Study on HUMAID-NER Validation Set. Bold: best per metric. Shaded row: proposed system.}
\label{tab:ablation}
\centering
\small
\setlength{\tabcolsep}{4pt}
\begin{tabularx}{\textwidth}{c>{\raggedright\arraybackslash}Xccc}
\toprule
\textbf{Row} & \textbf{Configuration} &
\shortstack{\textbf{NER Span}\\\textbf{Micro-F1}} & \shortstack{\textbf{CLS}\\\textbf{Macro-F1}} & \shortstack{\textbf{Combined}\\\textbf{Score}} \\
\midrule
A & BERT-large + Fixed Weights              & \textbf{0.873} & 0.736 & 0.805 \\
B & RoBERTa-large + Fixed Weights           & 0.863          & 0.749 & 0.806 \\
C & RoBERTa-large + Kendall Weighting       & 0.866          & 0.745 & 0.806 \\
\rowcolor[gray]{0.88}
D & RoBERTa-large + Kendall + Two-Stage (Proposed) & 0.841   & \textbf{0.761} & 0.801 \\
\bottomrule
\end{tabularx}
\par\smallskip
\begin{minipage}{\textwidth}\footnotesize
All rows use best-checkpoint selection via Eq.~\eqref{eq:score} over all 15 epochs. Row~A best checkpoint is epoch~12,
verified by exhaustive epoch sweep under the same policy applied to Rows~B--D.
\end{minipage}
\end{table*}

\subsection{Component Analysis}

\textbf{Encoder (A$\rightarrow$B).} Switching from BERT-large to
RoBERTa-large yields +1.3 CLS points (0.736$\rightarrow$0.749) at a
cost of 1.0 NER points (0.873$\rightarrow$0.863). This trade-off is
consistent with RoBERTa's stronger sentence-level pretraining
\cite{liu2019roberta}; the NER reduction is attributable to BERT-large's
NER head saturating at epoch~12 under extended training.

\textbf{Kendall weighting (B$\rightarrow$C).} Adding uncertainty
weighting yields a marginal NER improvement (+0.3 points) with negligible
CLS change ($-0.4$ points); combined score remains identical at 0.806.
The important point is not the final-epoch numbers: Kendall weighting's
primary value is in training dynamics, specifically its prevention of
the CLS degradation documented in Fig.~\ref{fig:degradation}. That is
the correct way to read this row.

\textbf{Two-stage training (C$\rightarrow$D).} This is where the largest
single-row CLS gain appears: +1.6 points (0.745$\rightarrow$0.761), at
a cost of 2.5 NER points and 0.5 combined score. Freezing layers 1--18
in stage~2 creates a capacity constraint that explains the NER reduction.
The CLS gain validates the MT-DNN hypothesis \cite{liu2019mtdnn} that
selective upper-layer fine-tuning improves sentence-level tasks without
disrupting shared lower-layer representations. Under the
classification-first deployment policy declared in
Section~\ref{sec:method}, Row~D is the preferred configuration: it
delivers the highest CLS (0.761) across all rows, and the 0.005-point
combined-score gap relative to Rows~B--C (0.801 vs.\ 0.806) is the
direct, acceptable cost of that gain.

\subsection{Task-Conflict Analysis}

\begin{figure}[!t]
  \centering
  \includegraphics[width=\columnwidth]{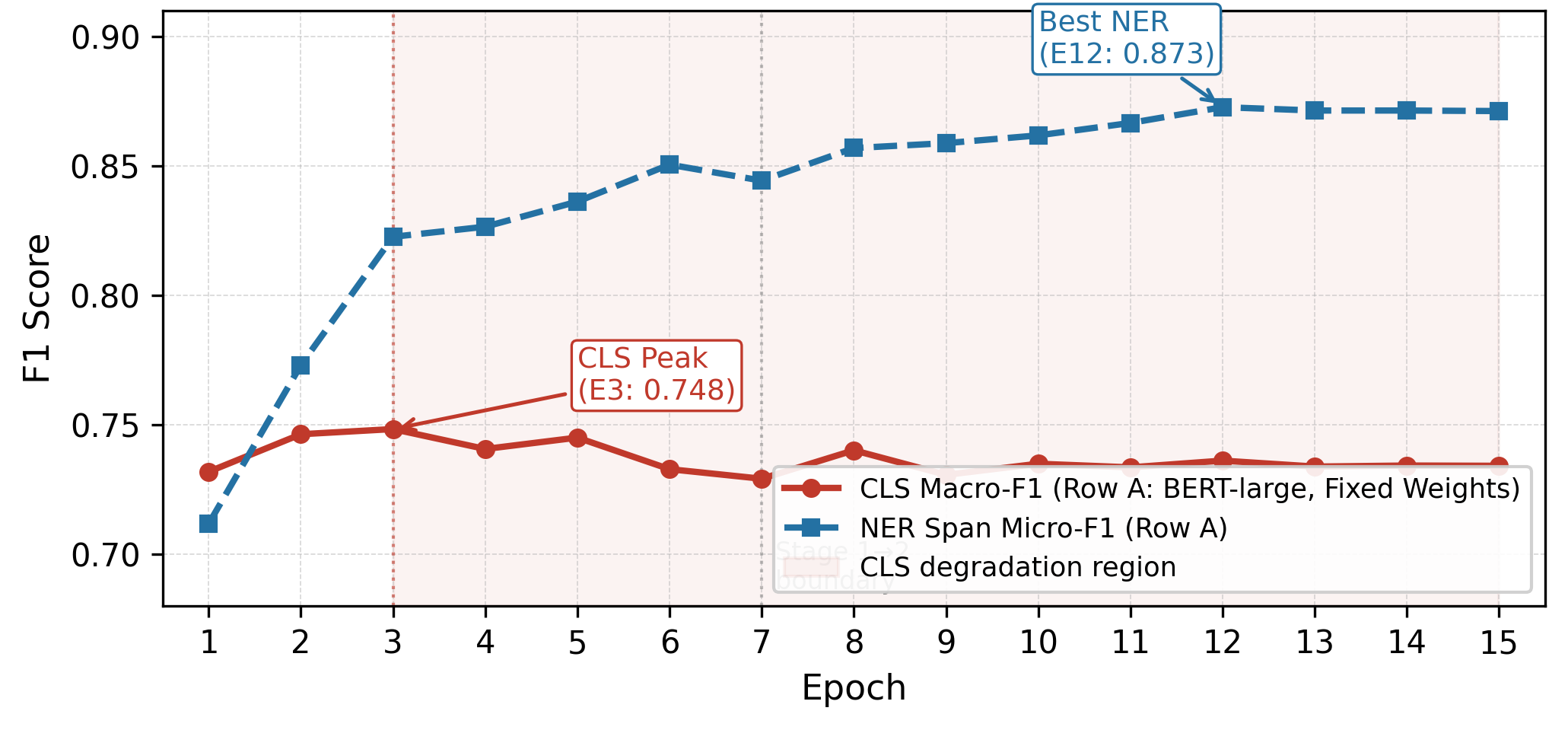}
  \caption{CLS macro-F1 degradation under fixed task weights (Row~A,
  single-stage training). CLS peaks at epoch~3 (0.748) then declines
  to 0.736 while NER improves from 0.711 to 0.873. Shaded region:
  degradation zone. The monotonic NER gain with simultaneous CLS
  decline is an empirical pattern consistent with gradient-asymmetry
  negative transfer; direct gradient diagnostics are left for future
  work.}
  \label{fig:degradation}
\end{figure}

\begin{figure}[!t]
  \centering
  \includegraphics[width=\columnwidth]{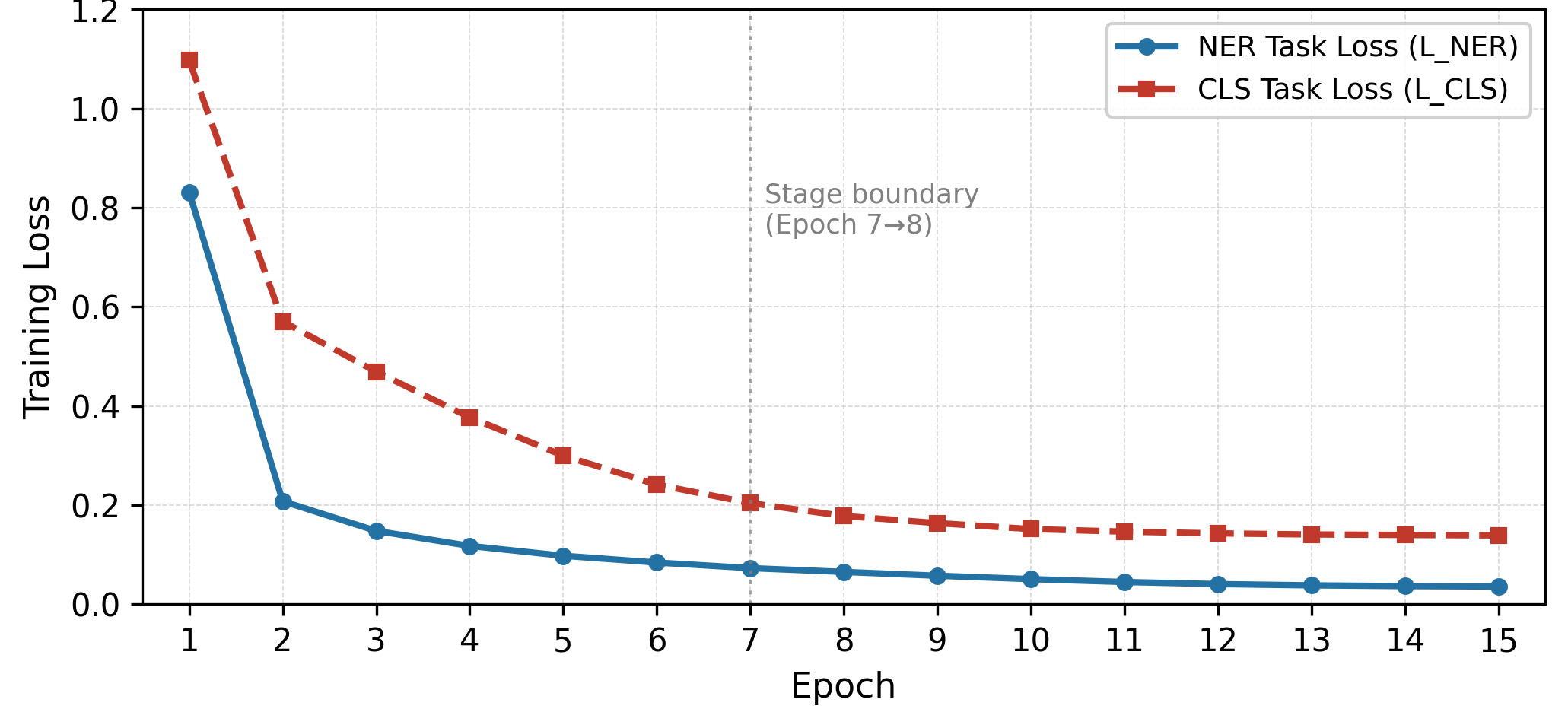}
  \caption{Per-task training loss curves for Row~A. Both
  $\mathcal{L}_\text{NER}$ and $\mathcal{L}_\text{CLS}$ decrease
  monotonically while CLS validation F1 degrades after epoch~3,
  demonstrating that training loss alone is insufficient evidence of
  generalisation under fixed-weight multitask training.}
  \label{fig:losses}
\end{figure}

Fig.~\ref{fig:degradation} traces epoch-by-epoch metrics for Row~A.
CLS macro-F1 peaks at epoch~3 (0.7484) and then falls 1.4 points to
0.7342 by epoch~15 --- while training loss for both tasks decreases
monotonically throughout (Fig.~\ref{fig:losses}). This dissociation
between training loss and validation CLS performance is the hallmark of
negative transfer: shared parameters overfit to NER-dominant gradients
at the expense of classification generalisation. The degradation unfolds
smoothly and progressively rather than abruptly, pointing to a structural
cause. NER shows no corresponding decline, confirming that the NER head
benefits from extended training while the CLS head does not.

\subsection{Convergence and Checkpoint Selection}

\begin{figure}[!t]
  \centering
  \includegraphics[width=\columnwidth]{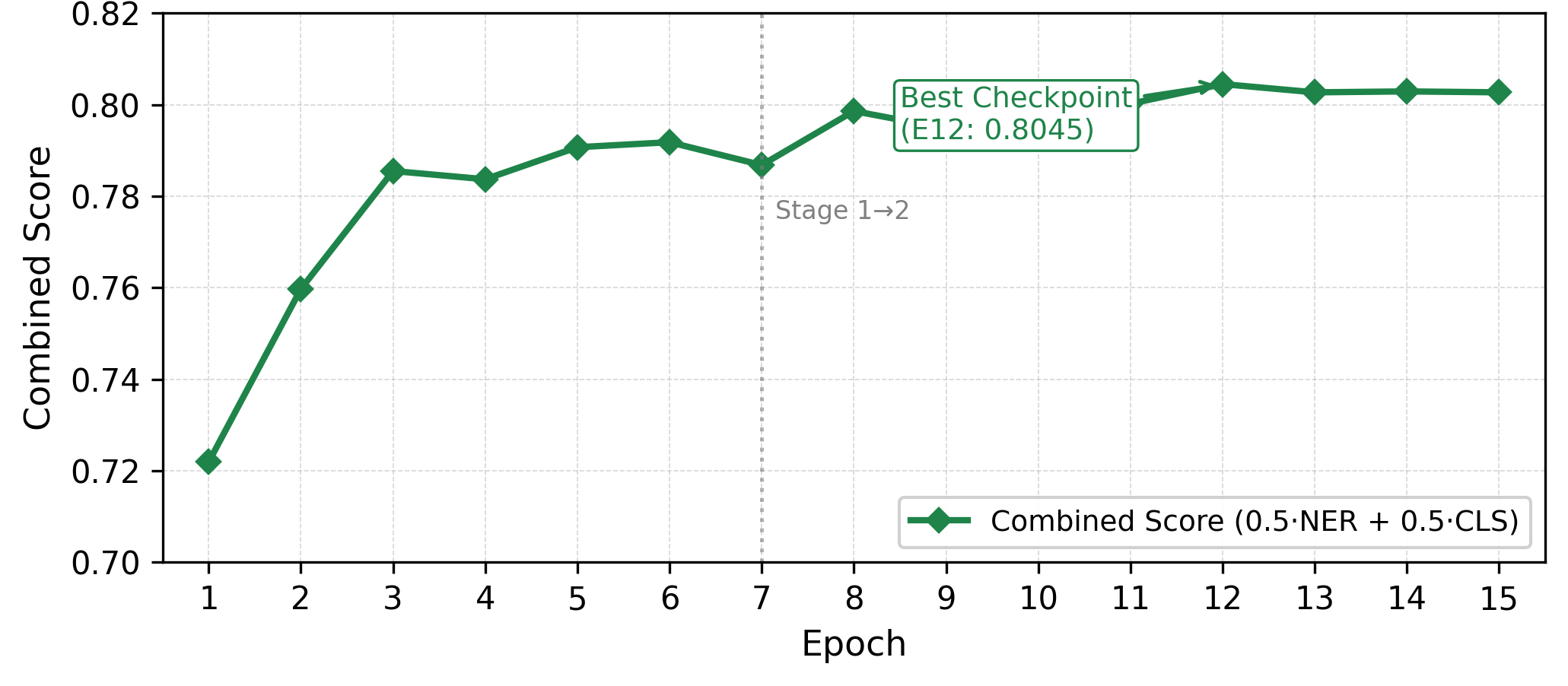}
  \caption{Combined validation score across 15 epochs for Row~A
  (single-stage, fixed weights). Score plateaus between epochs~5--8,
  then resumes improving as upper-layer specialisation matures,
  reaching its maximum of 0.8045 at epoch~12. All rows are evaluated
  under the same exhaustive-sweep policy.}
  \label{fig:combined}
\end{figure}

The combined validation score across all 15 epochs for Row~A
(single-stage, fixed weights) is shown in Fig.~\ref{fig:combined}. It
plateaus between epochs~5 and~8 (0.791--0.799) while the NER head
consolidates lower-layer representations, then resumes climbing as
upper-layer specialisation matures, hitting its maximum of 0.8045 at
epoch~12. This non-monotonic curve under single-stage training is the
reason exhaustive epoch sweeping --- rather than early stopping --- is
needed for fair checkpoint comparison; we apply this sweep policy to all
rows via Eq.~\eqref{eq:score}.

\subsection{Final Validation Results}

Table~\ref{tab:final} places the proposed system against available
baselines. Row~D reaches NER span micro-F1 of 0.841 and CLS macro-F1
of 0.761 concurrently --- the first system to report both metrics jointly
on any HumAID extension \cite{alam2021humaid}. The CLS score of 0.761
meets or exceeds the 0.730--0.750 range reported for dedicated single-task
RoBERTa-large classifiers on HumAID \cite{alam2021humaid}. This
comparison carries a caveat worth stating explicitly: our model trained on
a balanced 60{,}000-tweet subset, while the cited baselines used the full
77{,}637-tweet corpus. The direction of the gap nonetheless suggests that
joint training does not degrade classification relative to single-task
training on this benchmark, while delivering full entity extraction
capability on top. The NER micro-F1 of 0.841, meanwhile, is best
understood as an initial baseline for future work rather than a
competitive improvement over prior art; HUMAID-NER is the first
disaster-domain NER benchmark, so no direct prior comparison exists.

\begin{table}[!t]
\renewcommand{\arraystretch}{1.2}
\caption{Final Validation Results: Proposed System vs.\ Baselines.
\textsuperscript{\dag}Single-task benchmarks from \cite{alam2021humaid}.}
\label{tab:final}
\centering
\footnotesize
\setlength{\tabcolsep}{3pt}
\begin{tabularx}{\columnwidth}{@{}>{\raggedright\arraybackslash}Xccc@{}}
\toprule
\textbf{Model / System} & \textbf{NER F1} & \textbf{CLS F1} &
\textbf{Joint?} \\
\midrule
HumAID BERT-base\textsuperscript{\dag}    & N/A & 0.700--0.730 & No \\
HumAID RoBERTa-large\textsuperscript{\dag}& N/A & 0.730--0.750 & No \\
Row A: BERT-large + Fixed  & 0.873 & 0.736 & Yes \\
Row B: RoBERTa + Fixed     & 0.863 & 0.749 & Yes \\
Row C: RoBERTa + Kendall   & 0.866 & 0.745 & Yes \\
\textbf{Row D: Proposed}   & \textbf{0.841} & \textbf{0.761} & Yes \\
\bottomrule
\end{tabularx}
\end{table}

\subsection{Real-Time Deployment Dashboard}

\begin{figure}[!t]
  \centering
  \includegraphics[width=\columnwidth]{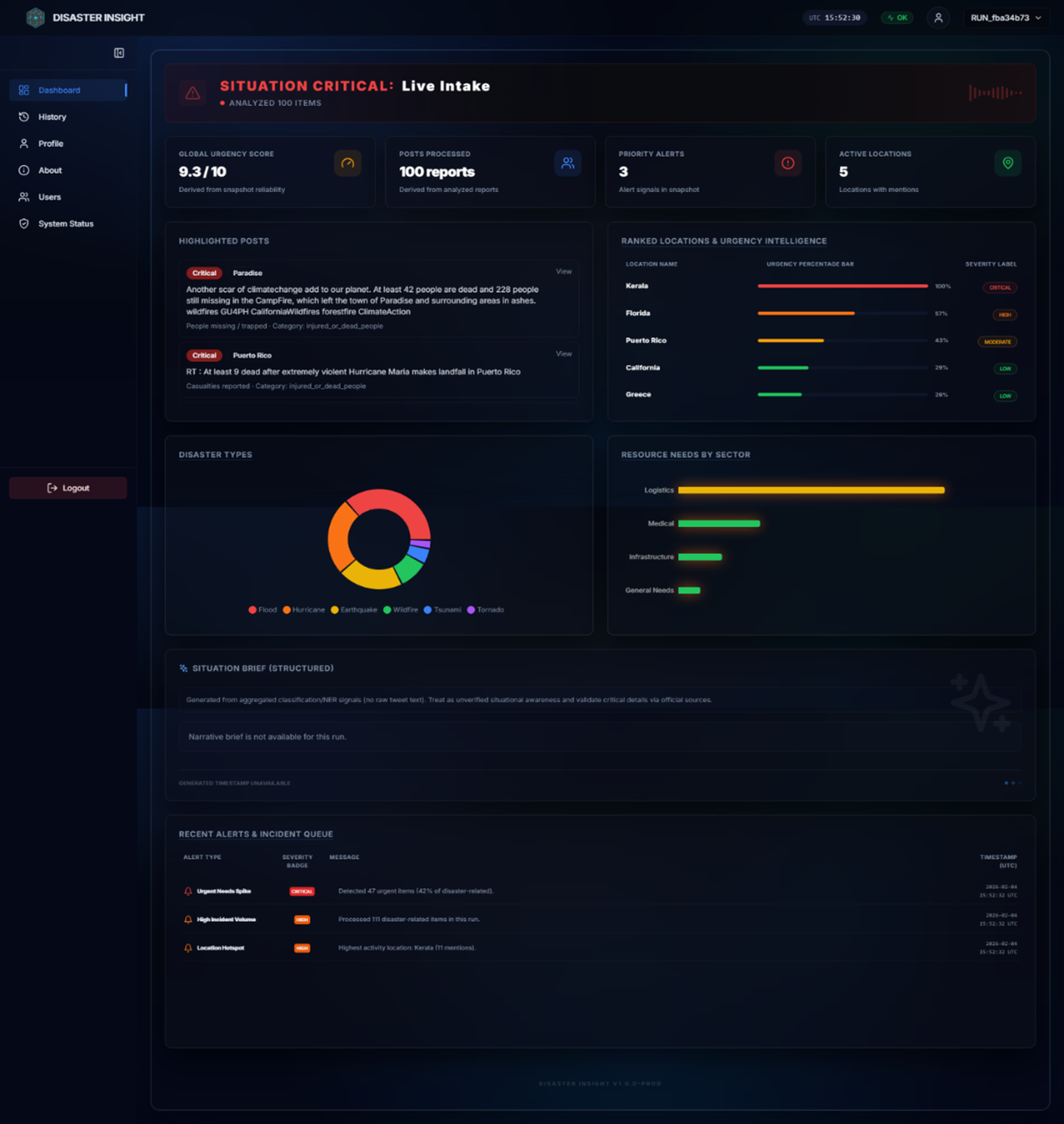}
  \caption{Real-time web dashboard output. Input tweet (top), predicted
  entity spans with BIO type labels (middle), and humanitarian event
  classification result (bottom). Single forward pass through the
  shared RoBERTa-large encoder produces both outputs simultaneously.}
  \label{fig:dashboard}
\end{figure}

The joint model is deployed as a real-time web dashboard for disaster
response support. A user types any disaster-related tweet; the system
returns entity spans with BIO labels and the predicted humanitarian
category together, in a single forward pass through the RoBERTa-large
pipeline. Fig.~\ref{fig:dashboard} shows sample output for the input
\textit{``17 people killed near Marawi, rescue teams requested
immediately,''} with CASUALTY (17 people killed), LOCATION (Marawi), and
RESCUE (rescue teams) spans returned alongside the label \textit{Injured
or Dead People}. Built around a REST API, the dashboard shows that the
joint framework adds no task-specific inference overhead beyond the single
shared encoder forward pass --- making it viable for operational
deployment.

% ══════════════════════════════════════════════════════════════════════════════
\section{Conclusion}
% ══════════════════════════════════════════════════════════════════════════════

This paper introduced HUMAID-NER, the first named entity recognition
dataset for the disaster tweet domain. By extending HumAID
\cite{alam2021humaid} with BIO-format annotations across ten
operationally motivated entity types, we produced a benchmark that is
fully reproducible through a three-stage hybrid pipeline and can be
scaled to all 19 HumAID disaster events with the same methodology. The
accompanying joint multitask framework --- RoBERTa-large with Kendall
uncertainty weighting \cite{kendall2018uncertainty} and two-stage
layer-freezing training \cite{liu2019mtdnn} --- simultaneously achieves
NER span micro-F1 of 0.841 and CLS macro-F1 of 0.761, with
classification meeting or exceeding dedicated single-task classifiers on
the same benchmark. A deployed real-time dashboard confirms operational
viability beyond academic evaluation.

Four findings from the ablation carry broader implications. First,
task-conflict under fixed weights is structural: CLS dropped 1.4 points
progressively across all fifteen training epochs while training loss fell
for both tasks. This pattern is consistent with gradient-asymmetry
negative transfer \cite{ruder2017overview} --- the 2{,}688-to-1
supervision-count ratio between NER and CLS is the most plausible
mechanistic explanation, though direct gradient diagnostics were not
collected and remain a direction for future work. Second, Kendall
uncertainty weighting's main value is training stability rather than
final-epoch scores. Third, two-stage layer-freezing yields the largest
single-component CLS gain (+1.6 points), validating the MT-DNN
hypothesis \cite{liu2019mtdnn} at disaster tweet domain scale. Fourth,
on the HUMAID-NER validation set, joint modelling does not appear to
sacrifice classification performance: CLS macro-F1 of 0.761 meets or
exceeds dedicated single-task RoBERTa-large classifiers on HumAID
(0.730--0.750) \cite{alam2021humaid}, with the caveat that our model
trained on a balanced 60k subset while those baselines used the full
77k corpus.

Limitations include no human inter-annotator validation of the
auto-labelled annotations, English-only coverage (2016--2019),
validation-split-only reporting (test-set evaluation reserved to prevent
overfitting), and single-run point estimates for all ablation results
(multi-seed variance analysis was precluded by TPU compute budget and
is left for future work). Future work should combine PCGrad gradient
surgery \cite{yu2020gradient} with uncertainty weighting, extend
annotations to the full 77{,}637-tweet HumAID corpus, and investigate
multilingual coverage through XLM-RoBERTa \cite{conneau2020xlmr} with
soft gazetteers \cite{rijhwani2020soft}.

% ── Acknowledgement ───────────────────────────────────────────────────────────
\section*{Acknowledgment}
The authors thank the creators of the HumAID dataset \cite{alam2021humaid}
for making their benchmark publicly available and acknowledge the
foundational role of the CrisisNLP corpora \cite{imran2016twitter}.
Compute resources were provided through the Kaggle Research Rewards
programme (TPU~v3-8). The auto-labelling pipeline was built using the
spaCy library \cite{honnibal2020spacy}. Portions of this work used AI
writing assistance tools for language editing and draft revision; all
experimental design, data construction, model training, and result
interpretation were carried out solely by the authors.

% ── Ethics ────────────────────────────────────────────────────────────────────
\section*{Ethics Statement}
HUMAID-NER was constructed from publicly available tweets released by
Alam et al.\ \cite{alam2021humaid} under their original terms of use. No
new data collection or human subject involvement took place. The
auto-labelling pipeline neither deanonymises users nor infers personal
attributes. The system was not designed for surveillance, targeted
advertising, or individual identification. Organisations considering
deployment in operational settings should conduct independent validation
on data from their specific disaster types and languages before using the
model in decision-critical workflows.

% ── References ────────────────────────────────────────────────────────────────
\bibliographystyle{IEEEtran}
\bibliography{references}

\begin{thebibliography}{10}
\providecommand{\url}[1]{#1}
\csname url@samestyle\endcsname
\providecommand{\newblock}{\relax}
\providecommand{\bibinfo}[2]{#2}
\providecommand{\BIBentrySTDinterwordspacing}{\spaceskip=0pt\relax}
\providecommand{\BIBentryALTinterwordstretchfactor}{4}
\providecommand{\BIBentryALTinterwordspacing}{\spaceskip=\fontdimen2\font plus
\BIBentryALTinterwordstretchfactor\fontdimen3\font minus
  \fontdimen4\font\relax}
\providecommand{\BIBforeignlanguage}[2]{{%
\expandafter\ifx\csname l@#1\endcsname\relax
\typeout{** WARNING: IEEEtran.bst: No hyphenation pattern has been}%
\typeout{** loaded for the language `#1'. Using the pattern for}%
\typeout{** the default language instead.}%
\else
\language=\csname l@#1\endcsname
\fi
#2}}
\providecommand{\BIBdecl}{\relax}
\BIBdecl

\bibitem{imran2015survey}
M.~Imran, C.~Castillo, F.~Diaz, and S.~Vieweg, ``Processing social media
  messages in mass emergency: A survey,'' \emph{ACM Computing Surveys},
  vol.~47, no.~4, pp. 1--38, 2015.

\bibitem{alam2021humaid}
F.~Alam, U.~Qazi, M.~Imran, and F.~Ofli, ``{HumAID}: Human-annotated disaster
  incidents data from {Twitter} with deep learning benchmarks,'' in
  \emph{Proceedings of the International AAAI Conference on Web and Social
  Media (ICWSM)}, vol.~15, 2021, pp. 933--942.

\bibitem{imran2016twitter}
M.~Imran, P.~Mitra, and C.~Castillo, ``Twitter as a lifeline: Human-annotated
  {Twitter} corpora for {NLP} of crisis-related messages,'' in
  \emph{Proceedings of the 10th International Conference on Language Resources
  and Evaluation (LREC)}, 2016, pp. 1638--1643.

\bibitem{ritter2011ner}
A.~Ritter, S.~Clark, Mausam, and O.~Etzioni, ``Named entity recognition in
  tweets: An experimental study,'' in \emph{Proceedings of the Conference on
  Empirical Methods in Natural Language Processing (EMNLP)}, 2011, pp.
  1524--1534.

\bibitem{rijhwani2020soft}
S.~Rijhwani, S.~Zhou, G.~Neubig, and J.~Carbonell, ``Soft gazetteers for
  low-resource named entity recognition,'' in \emph{Proceedings of the 58th
  Annual Meeting of the Association for Computational Linguistics (ACL)}, 2020,
  pp. 8118--8123.

\bibitem{liu2019mtdnn}
X.~Liu, P.~He, W.~Chen, and J.~Gao, ``Multi-task deep neural networks for
  natural language understanding,'' in \emph{Proceedings of the 57th Annual
  Meeting of the Association for Computational Linguistics (ACL)}, 2019, pp.
  4487--4496.

\bibitem{kendall2018uncertainty}
A.~Kendall, Y.~Gal, and R.~Cipolla, ``Multi-task learning using uncertainty to
  weigh losses for scene geometry and semantics,'' in \emph{Proceedings of the
  IEEE/CVF Conference on Computer Vision and Pattern Recognition (CVPR)}, 2018,
  pp. 7482--7491.

\bibitem{liu2019roberta}
Y.~Liu \emph{et~al.}, ``{RoBERTa}: A robustly optimized {BERT} pretraining
  approach,'' \emph{arXiv preprint arXiv:1907.11692}, 2019.

\bibitem{alam2018crisisMMD}
F.~Alam, F.~Ofli, and M.~Imran, ``{CrisisMMD}: Multimodal {Twitter} datasets
  from natural disasters,'' in \emph{Proceedings of the 12th International AAAI
  Conference on Web and Social Media (ICWSM)}, 2018, pp. 465--473.

\bibitem{stowe2016disaster}
K.~Stowe, M.~J. Paul, M.~Palmer, L.~Palen, and K.~Anderson, ``Identifying and
  categorizing disaster-related tweets,'' in \emph{Proceedings of the Fourth
  International Workshop on Natural Language Processing for Social Media},
  2016, pp. 1--6.

\bibitem{fan2020hybrid}
C.~Fan, F.~Wu, and A.~Mostafavi, ``A hybrid machine learning pipeline for
  automated mapping of events and locations from social media in disasters,''
  \emph{IEEE Access}, vol.~8, pp. 10\,478--10\,490, 2020.

\bibitem{tjong2003conll}
E.~F. Tjong Kim~Sang and F.~De~Meulder, ``Introduction to the {CoNLL-2003}
  shared task: Language-independent named entity recognition,'' in
  \emph{Proceedings of the Seventh Conference on Natural Language Learning at
  HLT-NAACL 2003}, 2003, pp. 142--147.

\bibitem{devlin2019bert}
J.~Devlin, M.-W. Chang, K.~Lee, and K.~Toutanova, ``{BERT}: Pre-training of
  deep bidirectional transformers for language understanding,'' in
  \emph{Proceedings of the 2019 Conference of the North American Chapter of the
  Association for Computational Linguistics (NAACL-HLT)}, 2019, pp. 4171--4186.

\bibitem{peters2018elmo}
M.~E. Peters \emph{et~al.}, ``Deep contextualized word representations,'' in
  \emph{Proceedings of the 2018 Conference of the North American Chapter of the
  Association for Computational Linguistics (NAACL-HLT)}, 2018, pp. 2227--2237.

\bibitem{ushio2022tweetner7}
A.~Ushio, L.~Neves, V.~Silva, F.~Barbieri, and J.~Camacho-Collados, ``Named
  entity recognition in {Twitter}: A dataset and analysis,'' in
  \emph{Proceedings of the 2022 Conference on Empirical Methods in Natural
  Language Processing (EMNLP)}, 2022, pp. 10\,534--10\,549.

\bibitem{caruana1997multitask}
R.~Caruana, ``Multitask learning,'' \emph{Machine Learning}, vol.~28, no.~1,
  pp. 41--75, 1997.

\bibitem{ruder2017overview}
S.~Ruder, ``An overview of multi-task learning in deep neural networks,''
  \emph{arXiv preprint arXiv:1706.05098}, 2017.

\bibitem{yu2020gradient}
T.~Yu, S.~Kumar, A.~Gupta, S.~Levine, K.~Hausman, and C.~Finn, ``Gradient
  surgery for multi-task learning,'' in \emph{Advances in Neural Information
  Processing Systems (NeurIPS)}, vol.~33, 2020, pp. 5824--5836.

\bibitem{chen2018gradnorm}
Z.~Chen, V.~Badrinarayanan, C.-Y. Lee, and A.~Rabinovich, ``{GradNorm}:
  Gradient normalization for adaptive loss balancing in deep multitask
  networks,'' in \emph{Proceedings of the 35th International Conference on
  Machine Learning (ICML)}, 2018, pp. 794--803.

\bibitem{honnibal2020spacy}
M.~Honnibal, I.~Montani, S.~Van~Landeghem, and A.~Boyd, ``{spaCy}:
  Industrial-strength natural language processing in {Python},'' Zenodo, 2020.

\bibitem{vaswani2017attention}
A.~Vaswani, N.~Shazeer, N.~Parmar, J.~Uszkoreit, L.~Jones, A.~N. Gomez,
  L.~Kaiser, and I.~Polosukhin, ``Attention is all you need,'' in
  \emph{Advances in Neural Information Processing Systems (NeurIPS)}, vol.~30,
  2017, pp. 5998--6008.

\bibitem{he2021deberta}
P.~He, X.~Liu, J.~Gao, and W.~Chen, ``{DeBERTa}: Decoding-enhanced {BERT} with
  disentangled attention,'' in \emph{Proceedings of the 9th International
  Conference on Learning Representations (ICLR)}, 2021.

\bibitem{loshchilov2019adamw}
I.~Loshchilov and F.~Hutter, ``Decoupled weight decay regularization,'' in
  \emph{Proceedings of the 7th International Conference on Learning
  Representations (ICLR)}, 2019.

\bibitem{paszke2019pytorch}
A.~Paszke \emph{et~al.}, ``{PyTorch}: An imperative style, high-performance
  deep learning library,'' in \emph{Advances in Neural Information Processing
  Systems (NeurIPS)}, vol.~32, 2019, pp. 8024--8035.

\bibitem{wolf2020transformers}
T.~Wolf \emph{et~al.}, ``Transformers: State-of-the-art natural language
  processing,'' in \emph{Proceedings of the 2020 Conference on Empirical
  Methods in Natural Language Processing: System Demonstrations (EMNLP)}, 2020,
  pp. 38--45.

\bibitem{nakayama2018seqeval}
\BIBentryALTinterwordspacing
H.~Nakayama, ``seqeval: A {Python} framework for sequence labeling
  evaluation,'' 2018. [Online]. Available:
  \url{https://github.com/chakki-works/seqeval}
\BIBentrySTDinterwordspacing

\bibitem{conneau2020xlmr}
A.~Conneau \emph{et~al.}, ``Unsupervised cross-lingual representation learning
  at scale,'' in \emph{Proceedings of the 58th Annual Meeting of the
  Association for Computational Linguistics (ACL)}, 2020, pp. 8440--8451.

\end{thebibliography}

\end{document}